\documentclass{article}
\usepackage{spconf,amsmath,graphicx}

\usepackage{enumitem}
\setlist{nosep, leftmargin=14pt}

\usepackage{mwe} 

\usepackage{amssymb,amsfonts}
\usepackage{booktabs}

\title{Self-Supervised Alignment Learning for Medical Image Segmentation}
\name{Haofeng Li$^{1,2,\star,\dagger}$ \qquad Yiming Ouyang$^{1,\star}$ \qquad Xiang Wan$^{1}$
\thanks{$\star$ denotes equal contributions. $\dagger$ denotes the corresponding author.}
}
\address{$^{1}$ Shenzhen Research Institute of Big Data, Shenzhen, China \\
     $^{2}$ The Chinese University of Hong Kong (Shenzhen), Shenzhen, China}
\begin{document}
%
\maketitle
\begin{abstract}
Recently, self-supervised learning (SSL) methods have been used in pre-training the segmentation models for 2D and 3D medical images. Most of these methods are based on reconstruction, contrastive learning and consistency regularization. However, the spatial correspondence of 2D slices from a 3D medical image has not been fully exploited. In this paper, we propose a novel self-supervised alignment learning framework to pre-train the neural network for medical image segmentation. The proposed framework consists of a new local alignment loss and a global positional loss. We observe that in the same 3D scan, two close 2D slices usually contain similar anatomic structures. Thus, the local alignment loss is proposed to make the pixel-level features of matched structures close to each other. Experimental results show that the proposed alignment learning is competitive with existing self-supervised pre-training approaches on CT and MRI datasets, under the setting of limited annotations. 
\end{abstract}
\section{Introduction}
\label{sec:intro} 
Lesion and organ segmentation~\cite{menze2014multimodal,bernard2018deep,xu2019whole,huang2022attentive} for medical images such as CT and MRI data are of great importance in disease diagnosis. For example, delineation of the left ventricular cavity, myocardium and right ventricle is essential for cardiac function analysis, and requires the automatic segmentation of Cardiac MRI~\cite{bernard2018deep}. As deep neural networks (DNNs) achieve success in generic visual recognition, they are also applied to medical image segmentation. 
However, most CT and MR images are three-dimensional (or even higher-dimensional) volume, which results in high labeling costs. 

To alleviate the issue of limited labels, self-supervised learning (SSL) algorithms provide a solution. Self-supervised pre-training for a DNN learns to capture generic representations from unlabeled data, and yields a robust initialization of model weights for the fine-tuning of downstream segmentation tasks. SSL with 2D images has been studied for years. Some of these methods degrade or destroy an unlabeled input image, and learn to extract features via reconstructing the original images~\cite{pathak2016context,fang2023corrupted,dong2022bootstrapped,yang2023mrm}. Some self-supervised approaches are based on contrastive learning~\cite{chen2020simple,misra2020self,peng2022crafting}, which trains a feature extractor by making the positive sample pair closer, and the negative pair further in a latent space. These pre-training methods can be applied to 2D slices of medical images, but they fail to exploit the inter-slice relationships. Recently, some SSL methods harvest more discriminative representations by using the prior knowledge and spatial structure of medical imaging data~\cite{zeng2021positional,peng2021self,ren2022local,li2023developing}, such as temporal similarity, patient ID and slice position. However, these methods pay less attentions to the pixel-level or region-level correspondence among slices. 

Inspired by the above observations, we propose to guide the learning of a DNN via locally matching medical image slices. Note that organs and tissues are 3D shapes, which are projected onto a series of continuous 2D slices. 
The region of the same tissue shows spatial coherence, and only expand or shrink slightly between two adjacent slices. Thus, for each image patch in a slice, it is very likely that from another neighboring slice we can find a matching patch of the same semantic. These two matching regions have similar anatomic structure but slightly different appearance. We argue that encouraging the encoder to embed the matching patches into similar features can help pre-train the model. However, labeling patch-level correspondence for two slices is even more expensive than labeling segmentation masks, and is prohibited. To solve the issue, we built a local alignment loss function. In the proposed loss, we compute dense pairwise similarity between the two slices, and maximize some of the higher similarity scores that likely correspond to the matching regions. To provide the global self-supervision, we further adopt a slice-level contrastive loss that takes the global features of nearby slices as a positive pair. The local alignment loss is integrated with the global one to build a complete pre-training framework for image segmentation. 

In this paper, we have three main contributions. First, we develop a novel local alignment loss that learns visual representations via matching local regions between slices. Second, we propose a self-supervised alignment learning framework for medical image segmentation, by combining the proposed alignment loss with a global positional loss. Third, experimental results show that the alignment loss can effectively improve the performance of downstream segmentation tasks.

\section{Method}
\label{sec:method}

\begin{figure}[!htb]
\centering
\includegraphics[width=0.48\textwidth]{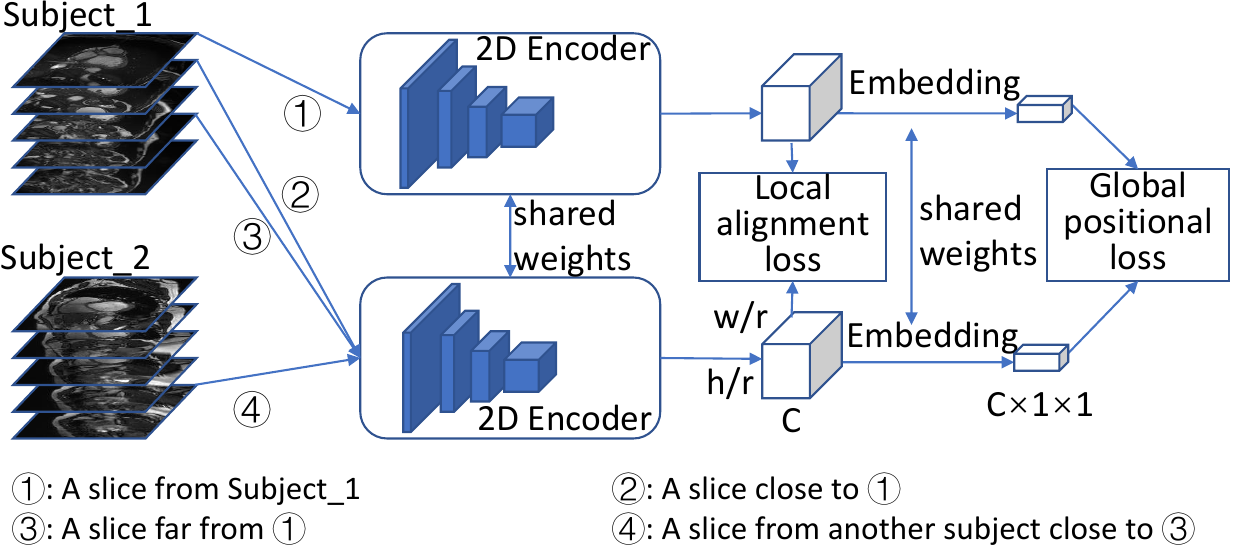}
\caption{The proposed Self-supervised Alignment Learning (SAL) framework. `Subject\_1\&2' denote two sequences of 2D slices. \textcircled{\scriptsize{1}}-\textcircled{\scriptsize{4}} represent different selections of slice pairs. The feature maps are divided into $r^2$ windows of size $\frac{h}{r}\times \frac{w}{r}$. }
\label{fig:framework}
\end{figure}

In this section, we propose a Self-supervised Alignment Learning (SAL) framework for initializing medical image segmentation models. SAL pre-trains the encoder of a 2D U-Net that can segment 2D slices of a 3D scan. As Fig.~\ref{fig:framework} shows, SAL has a novel local alignment (LA) loss and a global positional (GP) loss. The two losses are computed based on positive and negative slice pairs from the same subject or two different subjects. 

\subsection{Local Alignment Loss}
To learn pixel-level representations for image segmentation, we develop a local alignment loss. We observe that an organ or a lesion usually shows on multiple continuous slices in a 3D scan. Due to the spatial coherence of anatomical structures, there exists pixel-wise or patch-wise semantics correspondence among these 2D slices. We argue that if a model encodes the matching parts into close feature vectors, it is capable of depicting the semantics of the aligned regions. Due to lack of alignment labels, we resort to a self-supervised matching mechanism to formulate the LA loss.  

The inputs of LA loss are two $c \times h \times w$ feature maps ($c, h, w$ for channel, height, width), which are extracted for two slices that are near along the $z$ axis of the same 3D scan. $S_i$ and $S_j$ denote the two slices whose indices in the volume are $i$ and $j$, respectively. $F(\cdot)$ denotes the 2D encoder and $F(\cdot, s)$ returns the feature map of the $s^{th}$ scale ($s$ is set to 4) for an input slice. To compute cosine similarity, a linear embedding layer $E(\cdot)$ and a normalization operator $N(\cdot)$ are applied to these pixel-level feature maps. The normalized result is further reshaped to a $c\times hw$ matrix. For an input slice $S_i$, the above process is formulated as: $X_i = N(E(F(S_i, s)))$, where $X_i$ is the reshaped $c \times hw$ feature map. 

\begin{figure}[!ht]
\centering
\includegraphics[width=0.48\textwidth]{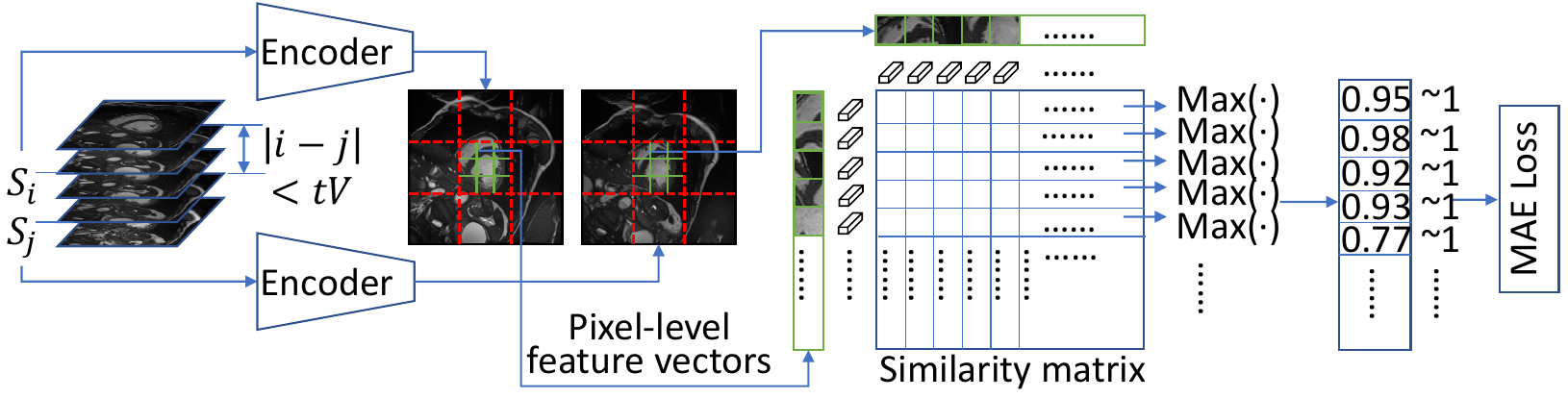}
\caption{The proposed Local Alignment loss function. $S_i$ and $S_j$ are two nearby slices in the same 3D image. Their feature maps are divided into windows by red dotted lines. 
`$\sim 1$' is to make the maximum similarity of each row close to 1, using MAE loss. $V$ is the number of slices and $t$ is the threshold deciding if two slices are used to compute the LA loss.} 
\label{fig:loss}
\end{figure}

For efficiency, the pixel-level similarity between $S_i$ and $S_j$ are estimated via matrix multiplication: $A = X_i^{T} X_j$, where $A$ is a $hw \times hw$ similarity matrix as shown in Fig.~\ref{fig:loss}. If the $k^{th}$ feature vector in $X_i$ has a matching vector in $X_j$, then the similarity between these two matching vectors should be the highest in the $k^{th}$ row of $A$ and close to 1. In Fig.~\ref{fig:loss}, a feature vector in $X_i$ corresponds to a small image patch in $S_i$. If $S_i$ and $S_j$ are visually close, then most image patches in $S_i$ have a matching one in $S_j$. Thus, we obtain the highest score in each row of $A$, and encourage the score to approximate to 1 using MAE loss. The LA loss is formulated as Eq.~(\ref{local_alignment_loss}): 
\begin{align}\label{local_alignment_loss}
    L^{align}(S_i, S_j) =& \textit{MAE} (max(X_i^T X_j, axis=0), 1),\notag \\
    &|i-j| < t \cdot V,
\end{align} 
where $V$ is the number of slices in the volume that contains $S_i$ and $S_j$. $t$ is a threshold in the range of $[0, 1]$, determining whether the two slices are close enough to compute the LA loss. $max(\cdot, axis=0)$ returns the maximum in each row of the input, producing a vector in Eq.~(\ref{local_alignment_loss}). $\textit{MAE}(\cdot, 1)$ computes the MAE loss between 1 and each element in the input vector/matrix, and returns the average of the element-wise losses. 

\subsection{Window-based Local Alignment loss}\label{sec:win_LA}
Calculating dense pairwise similarity across two feature maps is inefficiency, which brings a challenge for the LA loss. We consider that the visual contents slightly change across two nearby slices. The matching parts between two slices usually have close coordinates in the dimensions of height and width. For alignment it is unnecessary to traverse the whole $h\times w$ plane but only needs to go through a small window surrounding the query patch. Thus, we implement a variant of LA loss, Window-based Local Alignment (WLA) loss. The WLA loss splits the $h\times w$ plane of feature maps into $r^2$ windows, as shown in the red dotted lines in Fig.~\ref{fig:loss}. The alignment only happens within the two windows that have the same indices in the $h\times w$ plane. The shape of a window is $\frac{h}{r}\times \frac{w}{r}$.

For the naive LA loss, the time and space complexities are $(hw)^2 c$ and $h^2w^2$, respectively. For the WLA loss, the time complexity is $r^2 \times (\frac{h}{r} \frac{w}{r})^2 \times c = \frac{(hw)^2 c}{r^2}$, while the space complexity is $\frac{h^2w^2}{r^2}$. Therefore, the WLA loss can reduce the computing costs. We set the window sizes $\frac{h}{r}$ and $\frac{w}{r}$ to $\omega$. In Sec~\ref{sec:abla}, we discuss how to determine $\omega$ and how it affects the performance. Since the vanilla LA loss is not feasible, we use the WLA loss in practice. 

\subsection{Global Positional Loss}
To achieve slice-level representation learning, we implement a global positional (GP) loss to work with the local one. In typical contrastive learning, two views augmented from different source images (slices) are regarded as a negative pair. However, the definition of negative samples may be unsuitable for the 2D slices from 3D medical images. We note that for two different 3D volumes, their spatial distributions of organs and tissues are quite similar. For two slices that are from different scans but have close relative positions, it is very likely that they contain similar contents. Thus, if the position distance between slices is smaller than a threshold, then they should be regarded as positive pair~\cite{zeng2021positional}. The relative position of a slice can be computed via dividing its index by the total number of slices in the $z$ axis. The threshold is denoted as $t$, and is also used in LA loss. 

To calculate the GP loss, we randomly sampled $n$ slices during each training iteration. Data augmentation is applied to each sample to synthesize two views, which produces $2n$ samples. The GP loss of the $i^{th}$ sample is Eq.~(\ref{eq:pos}):
\begin{equation}\label{eq:pos}
L_i^{pos} = - \frac{1}{|P_i|} \sum_{j\in P_i} log \frac{Sim(f_i, f_j)}{\sum\nolimits_{q=1}^{2n} \mathbb{I}(i \neq q) Sim(f_i,f_q)}, 
\end{equation}
where $P_i$ is the set of positive samples of the $i^{th}$ slice, $Sim(\cdot,\cdot)$ is a similarity function used in ~\cite{zeng2021positional,chen2020simple}, $\mathbb{I}(\cdot)$ denotes the indicator function, $f_i$ and $f_j$ are two global feature vectors. To compute $f_i$, the feature map of the $i^{th}$ sample is first extracted by the encoder, and then processed by a global average pooling layer and a linear projection layer. 

\subsection{The Overall Loss Function}
This section describes how to aggregate the above two losses into a feature learning framework. Note that the definition of positive samples in the LA loss is different from that in the GP loss, and the LA loss requires that two slices in a positive pair are from the same volume. At the beginning of each training iteration, we first sample two 3D scans of different subjects, then randomly select $n/2$ slices from each sampled scan, and synthesize $2n$ samples using data augmentation. All pairs in these $2n$ samples are used to calculate the GP loss, while the LA loss only utilizes the positive pairs. The overall loss is defined as Eq.~(\ref{eq:overall_loss}): 
\begin{equation} \label{eq:overall_loss}
	L = \frac{1}{2n} \sum_{i=1}^{2n} L_i^{pos} + \lambda \cdot \frac{1}{|P^A|} \sum_{(S_i,S_j)\in P^A} L^{align}(S_i, S_j), 
\end{equation}
where $P^A$ is the set of positive pairs of slices for the LA loss. $\lambda$ is set to $1$ by default. We study the effect of different values of $\lambda$ in the experiment section.

\section{Experiments}
\label{sec:exper}

\subsection{Implementation details}
We adopt two public medical image segmentation datasets, ACDC~\cite{bernard2018deep} and CHD~\cite{xu2019whole}. ACDC is a cardiac MRI image dataset that consists of 3D MR images of 100 patients. For cross validation, the scans of 80 and 20 patients are used for training and validation, respectively. 
%
CHD includes 68 3D cardiac CT images. For cross validation, 54 subjects are adopted for training, while 14 subjects are for validation.  


Evaluating a pre-training method on a dataset has two stages, pre-training and fine-tuning. In the first stage, we utilize the pre-training method and the whole dataset without labels to train the encoder of a 2D U-Net. In the second stage, we fine-tune and evaluate the U-Net on the dataset with labels, in a 5-fold cross-validation manner. At the beginning of fine-tuning, the pre-trained encoder is used to initialize the 2D U-Net. We report the mean and standard deviation of the results obtained by the 5-fold cross validation. Since we aim at improving label-efficient image segmentation with SSL pre-training, we report the results of using limited annotations in the fine-tuning stage. In the label-limited setting, at each fold only the images of $M$ randomly selected subjects and their annotations are used for fine-tuning. For ACDC dataset, $M$ is set to 2, 6, 10, and 80. For CHD dataset, $M$ is set to 2, 6, 15, and 51. At each fold, the training samples of different pre-training methods are exactly the same. Dice similarity coefficient (DSC) is adopted as the evalutation metric.

\begin{table}[!t]
\caption{Comparison with the existing pre-training methods on the ACDC dataset. $M$ denotes the number of subjects used for fine-tuning. `Random' means training from scratch. }
\label{tab:sota_acdc}
\centering
\footnotesize
\begin{tabular}{lllll}
\toprule
Method & \multicolumn{4}{c}{Mean DSC (Std)} \\
\cmidrule(r){2-5}
&  $M=2$ & $M=6$ & $M=10$ & $M=80$ \\
\midrule
Random      &  0.588(.07) & 0.782(.03) & 0.840(.03) & 0.928(.00) \\
Rotation~\cite{gidaris2018unsupervised}    &  0.572(.08) & 0.809(.03) & 0.868(.02) & 0.925(.00) \\
PIRL~\cite{misra2020self}        &  0.492(.03) & 0.823(.04) & 0.865(.01) & 0.927(.00) \\
SimCLR~\cite{chen2020simple}      &  0.352(.06) & 0.725(.08) & 0.824(.04) & 0.927(.00) \\
GCL~\cite{chaitanya2020contrastive}         &  0.636(.05) & 0.803(.04) & 0.872(.01) & 0.927(.01) \\
PCL~\cite{zeng2021positional}        &  \underline{0.671}(.06) & \underline{0.850}(.01) & \underline{0.885}(.01) & \underline{0.929}(.00) \\
CCrop~\cite{peng2022crafting}       &  0.262(.07) & 0.729(.06) & 0.853(.02) & 0.926(.00) \\
Ours        &  \textbf{0.723}(.04) & \textbf{0.869}(.02) & \textbf{0.890}(.01) & \textbf{0.929}(.00) \\
\bottomrule
\end{tabular}
\end{table}

\begin{table}[!ht]
\caption{Comparison between the existing pre-training methods and ours on the CHD dataset.}
\label{table:chd}
\centering
\footnotesize
\begin{tabular}{lllll}
\toprule
Method & \multicolumn{4}{c}{Mean DSC (Std)}                   \\
\cmidrule(r){2-5}
&  $M=2$ & $M=6$ & $M=15$ & $M=51$ \\
\midrule
Random      &  0.184(.06) & 0.508(.06) & 0.627(.05) & 0.754(.02) \\
Rotation~\cite{gidaris2018unsupervised}    &  0.171(.06) & 0.488(.07) & 0.625(.04) & 0.749(.03) \\
PIRL~\cite{misra2020self}        &  0.196(.07) & 0.504(.08) & 0.658(.03) & 0.761(.03) \\
SimCLR~\cite{chen2020simple}      &  0.192(.06) & 0.515(.06) & 0.631(.05) & 0.756(.03) \\
GCL~\cite{chaitanya2020contrastive}        &  0.321(.05) & 0.549(.06) & 0.676(.04) & 0.760(.04) \\
PCL~\cite{zeng2021positional}        &  \underline{0.338}(.04) & \underline{0.566}(.05) & 0.685(.04) & \underline{0.771}(.03) \\
CCrop~\cite{peng2022crafting}       &  0.256(.04) & 0.548(.06) & \underline{0.690}(.05) & 0.769(.03) \\
Ours        &  \textbf{0.348}(.04) & \textbf{0.572}(.06) & \textbf{0.693}(.04) & \textbf{0.772}(.03) \\
\bottomrule
\end{tabular}
\end{table}

\subsection{Comparison with the State-of-the-art Methods}

To show the effectiveness, our proposed method is compared with the existing SSL pre-training approaches, Rotation~\cite{gidaris2018unsupervised}, PIRL~\cite{misra2020self}, SimCLR~\cite{chen2020simple}, GCL~\cite{chaitanya2020contrastive}, PCL~\cite{zeng2021positional} and CCrop~\cite{peng2022crafting}. Table~\ref{tab:sota_acdc} shows the results on ACDC dataset. $M$ refers to the number of subjects adopted in fine-tuning. In Table~\ref{tab:sota_acdc}, as $M$ increases from 2 to 80, the mean Dice of each method continuously rises. When $M$ is low, increasing labeled samples significantly improves the mean Dice by $14.6\%$-$37.3\%$. When $M$ is high (from 10 to 80), the great increase of samples only brings smaller improvements of $3.9\%$-$10.3\%$ mean Dice. The `Random' strategy is to randomly initialize model weights without any pre-training. It can be seen that when using all training samples ($M=80$), the mean Dice of the Random and most pre-training methods (including ours) are very close (92.7\%-92.8\%). It indicates that the SSL-based pre-training usually shows the effectiveness when training samples are not completely labeled. For $M=2$, our method outperforms the second-best PCL by $5.2\%$ mean Dice. For $M=6$ and $M=10$, our framework surpasses the second best by $0.5\%$-$1.9\%$. These results suggest that the proposed SAL framework is competitive with existing pre-training approaches. 

Table~\ref{table:chd} is the comparison on a CT image dataset, CHD. Since the results reported in ~\cite{zeng2021positional} are not reproduced, we re-implement and evaluate GCL and PCL. Table 2 shows the similar phenomena to Table 1. As $M$ increases from 2 to 51, the mean Dice of all methods first rise greatly and then grow slowly. When setting $M$ to $2,6$, our method surpasses the second best method PCL by $1.0\%$, $0.6\%$, respectively. The results on two datasets verify that the SAL has strong generalization ability and can well adapt to both CT and MR images.

\subsection{Ablation Study and Hyper-parameter Analysis}
\label{sec:abla}

In this section, we study two important hyper-parameters of our method, the loss weight $\lambda$ and the window size $\omega$.  After setting the two hyper-parameters to a series of values, we report the mean and standard deviation of Dice scores from 5-fold cross validation, on the ACDC dataset with $M=6$. 

\textbf{Loss weight and LA loss.} $\lambda$ is to make a balance between the LA and the GP losses. In Table~\ref{tab:loss_weight}, we set $\lambda$ to $0, 1, 5, 40, 80$. When $\lambda$ increases from 0 to 80, the mean Dice boosts from 85.0\% to 86.9\%, and then gradually decreases to 85.3\%. $\lambda=0$ means removing the LA loss from our method. Comparing $\lambda=1$ to $\lambda=0$ shows that the LA loss effectively improves the mean Dice by 2\%. Setting $\lambda=80$ may pay too less attentions to the GP loss and leads to a drop of 1.6\%. That is to say, the LA and GP losses are complementary, and it is reasonable to integrate them into our SAL framework. 

\begin{table}[!t]
	\caption{Investigation of the weight of the proposed Local Alignment loss. $\lambda$ denotes the weight of LA loss.} 
\label{tab:loss_weight}
\centering
\footnotesize
\begin{tabular}{lllll}
	\toprule
	\multicolumn{5}{c}{Mean DSC (Std) on ACDC Dataset with $M = 6$} \\
	\cmidrule(r){1-5}
	$\lambda=0$ &  $\lambda=1$ & $\lambda=5$ & $\lambda=40$ & $\lambda=80$ \\
	\midrule
	0.850(.01)      &  0.869(.02) & 0.868(.02) & 0.868(.01) & 0.853(.02) \\
	\bottomrule
\end{tabular}
\end{table}

\begin{table}[!t]
	\caption{Investigation of the window size of the proposed Local Alignment loss. $\omega$ denotes the window size. 
		It can be seen that the proposed method is insensitive to the change of $\omega$.}
	\label{tab:window_size}
	\centering
	\footnotesize
	\begin{tabular}{lllll}
		\toprule
		\multicolumn{5}{c}{Mean DSC (Std) on ACDC Dataset with $M = 6$} \\
		\cmidrule(r){1-5}
		Ours w/o LA loss & $\omega=2$ &  $\omega=3$ & $\omega=4$ & $\omega=6$ \\
		\midrule
		0.850(.01)      &  0.869(.02) & 0.868(.01) & 0.869(.02) & 0.867(.01) \\
		\bottomrule
	\end{tabular}
\end{table}

\textbf{Window size.} $\omega$ denotes the window size that controls the computation cost and the area in which the pixels from two feature maps are aligned. As Figure~\ref{fig:loss} displays, the feature maps of two slices are evenly partitioned into windows when computing the similarity matrix in LA loss. Each window contains $\omega^2$ feature vectors. If $\omega$ is larger, then the computation cost is higher, as discussed in Sec~\ref{sec:win_LA}. Due to the limit of computing resource, $\lambda$ is not set to a value larger than 6. In Table~\ref{tab:window_size}, as $\omega$ grows from $2$ to $6$, the mean Dice score does not change much but only shows a slight decrease form 86.9\% to 86.7\%. It is noteworthy that no matter which value $\omega$ is set to, our method brings a notable improvement of $1.7\%$-$1.9\%$ mean Dice, compared to the one without LA loss. It suggests that our method is robust to the change of window size.

\section{Conclusion}
In this paper, we propose a novel Self-supervised Alignment Learning (SAL) framework that aims to improve the slice-based 3D medical image segmentation by pre-training. The proposed SAL framework learns the parameters of an image segmentation model at both local pixel level and global slice level. For pixel-level learning, we exploit the dense semantic correspondence between closely located slices in a 3D volume, and develop a new Local Alignment (LA) loss that guides the model training by self-supervised feature matching. Furthermore, a global positional (GP) loss, which views the slices of close relative positions as positive pair, is integrate with LA loss to build the overall SAL framework. Extensive experiments show that our proposed pre-training framework obtains the state-of-the-art results on both CT and MRI datasets, especially with a small number of labels for fine-tuning. Besides, the results verify the effectiveness of LA loss in the framework.

\section{Compliance with ethical standards}
\label{sec:ethics}
The research was conducted using human subject data made available in open access. Ethical approval was not required as confirmed by the license attached with the open access data.

\section{Acknowledgments}
This work is supported in part by Chinese Key-Area Research and Development Program of Guangdong Province (2020B0101350001), the National Natural Science Foundation of China (No.62102267), and the Guangdong Basic and Applied Basic Research Foundation (2023A1515011464).


\bibliographystyle{IEEEbib}
\bibliography{strings,refs}

\end{document}